\documentclass[lettersize, 10 pt, journal, twoside]{ieeeconf}

\usepackage{amsmath,amsfonts}
\usepackage{algorithmicx}
\usepackage{algorithm}
\usepackage{array}
\usepackage[caption=false,font=normalsize,labelfont=sf,textfont=sf]{subfig}
\usepackage{textcomp}
\usepackage{stfloats}
\usepackage{url}
\usepackage{verbatim}
\usepackage{graphicx}
\usepackage{cite}
\usepackage{graphics}
\usepackage{bm}
\usepackage{multirow}
\usepackage[algo2e]{algorithm2e}
\usepackage{makecell}
\usepackage{amssymb}
\usepackage{booktabs}
\usepackage[margin=2cm]{geometry}
\usepackage{xcolor}
\usepackage{pifont}
\usepackage[compatible]{algpseudocode}
\usepackage{threeparttable}
\usepackage{xurl}

\IEEEoverridecommandlockouts
\begin{document}

\title{\LARGE \bf
Safe Navigation in Unknown and Cluttered Environments via Direction-Aware Convex Free-Region Generation}

\author{Zhicheng Song$^{1}$, Yongjian Li$^{1}$, Kai Chen$^{1}$, Yulin Li$^{2}$, Fan Shi$^{2}$, and Jun Ma$^{1}$ 
\thanks{$^{1}$Zhicheng Song, Yongjian Li, Kai Chen, and Jun Ma are with the Robotics and Autonomous Systems Thrust, The Hong Kong University of Science and Technology (Guangzhou), Guangzhou, China (e-mail: zsong469@connect.hkust-gz.edu.cn; yli125@connect.hkust-gz.edu.cn; kchen916@connect.hkust-gz.edu.cn; jun.ma@ust.hk)}
\thanks{$^{2}$Yulin Li and Fan Shi are with the Department of Electrical and Computer Engineering, National University of Singapore, Singapore (e-mail:yline@nus.edu.sg; fan.shi@nus.edu.sg)}
}

\maketitle

\begin{abstract}
Convex free regions provide a structured and optimization-friendly representation of collision-free space for robot navigation in unknown and cluttered environments. 
However, existing methods typically enlarge local collision-free regions mainly according to surrounding obstacle geometry. 
In cluttered environments, such strategies may fail to generate regions that both accommodate robot geometry and preserve traversable extension along candidate motion directions, thereby limiting downstream traversal, especially in narrow passages.
Even when such a region is available, safe motion generation remains challenging, because safety checking at discretized trajectory samples does not guarantee continuously collision-free motion when robot geometry is modeled explicitly. 
To address these issues, we propose a navigation framework that jointly incorporates candidate motion directions and robot geometry into convex free-region generation, and achieves continuously collision-free motion through continuous-safe trajectory generation. 
Within each region, the framework performs geometry-aware target pose selection and trajectory generation, together with Lipschitz-based continuous safety certification and local refinement. 
The resulting free regions and candidate motions are maintained in a region-based graph to support incremental planning. 
Quantitative results in cluttered 2D navigation scenarios show that the proposed method generates free regions better aligned with downstream traversal and enables reliable collision-free navigation, while additional 3D and real-world experiments on a quadrupedal robot and a UAV demonstrate the extensibility and practical applicability of the framework.
The open-source project can be found at \url{https://github.com/ZhichengSong6/FRGraph}.
\end{abstract}

\section{INTRODUCTION}
 
Collision-free navigation in cluttered and unknown environments requires a free-space representation that is both structured and amenable to downstream motion generation. 
Among existing representations, convex free-space decomposition is particularly attractive because it converts discrete obstacle observations into structured collision-free regions that can be used directly for trajectory generation and online replanning~\cite{SikangLiu, RussDecompose, faster}. 
Such regions also provide a convenient geometric form for explicitly reasoning about robot geometry in both planar and spatial navigation settings~\cite{FRTree, ylSOS, ylLocalReactive}.

\begin{figure}[tp]
    \centering
    \includegraphics[width = 0.47\textwidth]{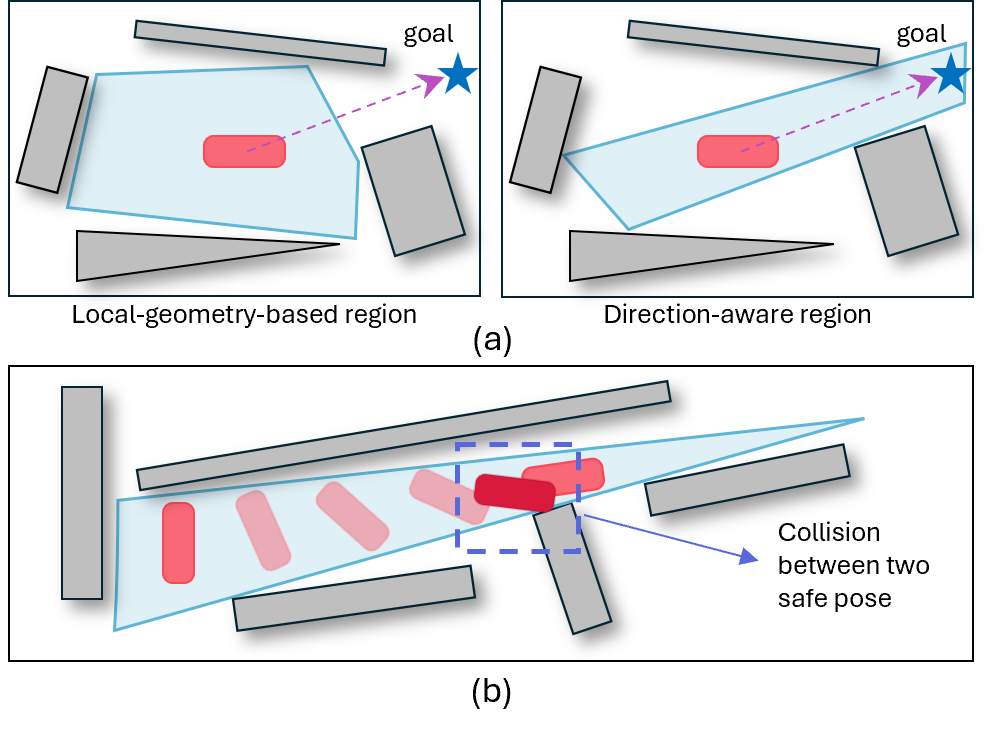} 
    \vspace{-0.3cm}
    \caption{Motivation of the proposed framework. Gray polygons denote obstacles, the blue polygon denotes the convex free region, the red rectangle denotes the robot, and the purple dashed arrow indicates the desired motion direction.
    (a) If the free region is generated solely according to the surrounding obstacle configuration, the resulting collision-free region may provide limited extension along the desired direction. This motivates direction-aware free-region generation that better preserves traversable space for forward motion.
    (b) With explicit robot geometry, a collision may occur between two safe sampled poses, motivating continuous-time safety verification and refinement.}
    \label{fig:motivation}
    \vspace{-0.5cm}
\end{figure}

However, for navigation, the quality of a convex free region is determined not only by collision avoidance, but also by whether it accommodates robot geometry and preserves traversable space along candidate motion directions. 
Existing convex free-region generation methods mainly construct regions according to local obstacle geometry~\cite{SikangLiu, GaoFeiFreeSpace, FRTree}. 
Although such methods can produce geometrically valid collision-free regions, they are primarily designed to enlarge local free space and do not explicitly incorporate candidate motion directions or robot geometry into region construction. 
As a result, in cluttered scenes with narrow passages, the generated region may be poorly suited to downstream traversal even when it is collision-free, as illustrated in Fig.~\ref{fig:motivation}(a). 
This observation motivates a free-region generation strategy that jointly accounts for candidate motion directions and robot geometry.

After a convex free region has been constructed, generating safe motion within that region remains challenging when robot geometry is modeled explicitly.
Many navigation systems simplify robot geometry through point approximation or obstacle inflation~\cite{kai,sphere}, reducing safety checking to verifying a single curve~\cite{faster,kai,otte2016rrtx}. 
When robot geometry is modeled explicitly, however, collision-free sampled poses do not guarantee collision-free motion between them, as illustrated in Fig.~\ref{fig:motivation}(b). 
Nevertheless, many navigation methods based on convex free regions still verify safety only at discretized trajectory samples~\cite{FRTree,Navigationin2DClutteredScenes}, which may miss such inter-sample collisions.

To address these challenges, we propose a navigation framework that integrates direction-aware convex free-region generation and continuous-safe trajectory generation for robot navigation in unknown and cluttered environments.
Starting from local LiDAR observations, the framework extracts candidate navigation directions to construct convex regions that better preserve traversable space along those directions while accommodating robot geometry.
Within each region, it performs target pose selection, trajectory generation, and continuous safety verification under explicit robot-geometry constraints.
The resulting free regions and candidate motions are further maintained in a region-based graph for incremental planning.
In summary, the main contributions of this work are as follows:
\begin{itemize}
    \item We propose a direction-aware convex free-region generation method that incorporates candidate navigation directions into obstacle selection and hyperplane optimization through a direction-biased quadratic program (QP), producing free regions that better preserve traversable space along candidate directions while accommodating robot geometry in cluttered environments.

    \item We develop a continuous-safe trajectory generation method for convex-shaped robots inside convex free regions, combining Bézier trajectory parameterization with Lipschitz-based adaptive interval subdivision for continuous safety certification and local QP-based control-point refinement for iterative correction.
    
    \item We integrate the above components with a region-based navigation graph for incremental planning and recovery from locally unproductive branches, and validate the open-source framework through quantitative 2D experiments together with 3D and real-world demonstrations.
\end{itemize}
 
\section{RELATED WORK}

\subsection{Free-Space Representations for Navigation}

Existing navigation methods adopt a variety of collision-free space representations, including signed distance fields (SDFs)~\cite{ESDF_UniversalTO, ESDF-Robo-Centric}, implicit collision constraints based on separating hypersurfaces~\cite{hypersurface,syHypersurface}, and convex free-space decompositions~\cite{EllipsoidalRepresentation, Fast-Racing, FRTree,ylLocalReactive,ylSOS, Navigationin2DClutteredScenes, SikangLiu, GaoFeiFreeSpace, RussDecompose}. 
Among these, convex free-space decompositions are particularly attractive for online navigation because they provide structured geometric constraints that can be used directly for target selection, trajectory generation, and replanning.


Most existing convex free-region generation methods construct regions mainly according to local obstacle geometry. 
RILS~\cite{SikangLiu} does not guarantee that the generated region fully contains the robot body. 
Although a line-segment seed can approximately encourage region growth along a desired direction, it may produce overly slender and unhelpful regions in narrow spaces. 
FRTree~\cite{FRTree} mitigates this issue through overlapping line-segment seeds, but still does not guarantee full robot containment or fully avoid such slender regions. 
FIRI~\cite{GaoFeiFreeSpace} can generate regions that contain the robot geometry, but it does not explicitly consider the intended motion direction during region generation.
Table~\ref{tab:region_generation_comparison} summarizes these differences among representative convex free-region generation methods.

\begin{table}[tbp]
    \centering
    \caption{Comparison of convex free-region generation.}
    \label{tab:region_generation_comparison}
    \setlength{\tabcolsep}{5.0pt}
    \begin{tabular}{c c c}
        \toprule
        Method & 
        \makecell{Direction-Aware\\Region Generation} & 
        \makecell{Explicit Robot-Body\\Containment} \\
        \midrule
        RILS~\cite{SikangLiu}  & \textcolor{red}{\ding{55}} & \textcolor{red}{\ding{55}} \\
        FRTree~\cite{FRTree} & \textcolor{red}{\ding{55}} & \textcolor{red}{\ding{55}} \\
        FIRI~\cite{GaoFeiFreeSpace} & \textcolor{red}{\ding{55}} & \textcolor{green!60!black}{\ding{51}} \\
        Ours & \textcolor{green!60!black}{\ding{51}} & \textcolor{green!60!black}{\ding{51}} \\
        \bottomrule
    \end{tabular}
\vspace{-0.5cm}
\end{table}

\subsection{Navigation Direction Extraction}

Navigation directions can be extracted from local observations in different ways. 
Frontier-based methods guide the robot toward boundaries between explored and unexplored space~\cite{Frontier1997,3DFrontier,2024RALFrontier}, typically at the map level for high-level planning.

Another line of work extracts directions from local geometric openings in sensor observations. 
Gap-based approaches identify traversable directions from the spatial arrangement of obstacles and guide the robot toward openings likely to admit collision-free motion~\cite{gapbased2010,safergap,potentialgap}. 
Gap-based navigation has been studied mainly in planar settings, with relatively limited work in full 3D environments. 
Compared with frontier-based strategies, they provide local geometric cues that are more naturally suited to downstream free-region generation from local perception.

\subsection{Safe Trajectory Generation}

Trajectory generation methods differ substantially in how robot geometry is handled during collision checking. 
Many real-time planning methods simplify robot geometry during collision checking, so that safety evaluation reduces to verifying a single curve in free space~\cite{kai,sphere,faster,otte2016rrtx}.

When robot geometry is modeled explicitly, ensuring collision-free motion becomes more challenging. 
Even if the start and goal poses are feasible, and even if sampled states along the trajectory are collision-free, the robot may still violate safety constraints between adjacent samples, especially in cluttered environments with small geometric clearances. 
Although sampled checking is often effective in practice~\cite{FRTree, Navigationin2DClutteredScenes}, it does not provide a certificate of continuous-time safety when robot geometry is modeled explicitly.
Related continuous-time safety issues have been studied in prior work on convex collision checking, continuous-time collision avoidance, and continuous collision detection for polynomial trajectories~\cite{continuousCollisionCheck, merkt2019continuous, zhang2022generalized}.
\section{METHOD}

\begin{figure*}[t]
    \centering
    \includegraphics[width = 1.0\textwidth]{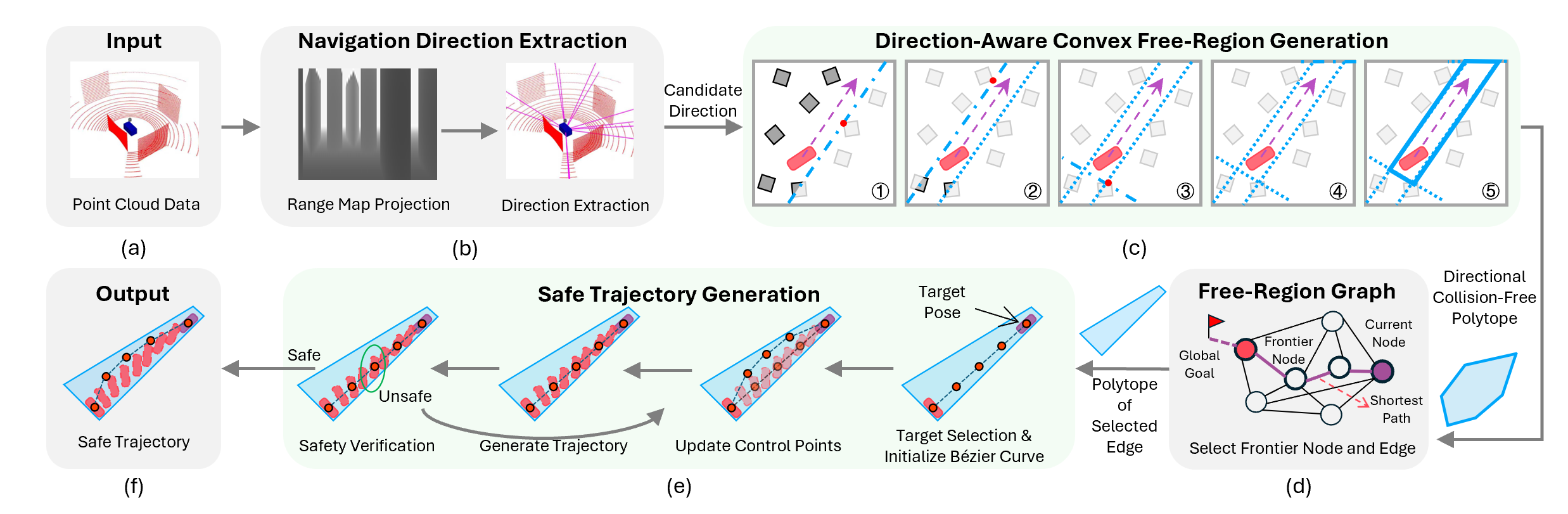}
    \caption{
    Overview of the proposed navigation framework. 
    From LiDAR point cloud observations, candidate navigation directions are extracted via range-map projection and gap-based direction extraction. 
    These directions are then used to generate direction-aware convex free regions that better support downstream robot traversal. 
    Within a selected free region, a target pose is determined and a B\'ezier trajectory is iteratively refined through continuous safety verification and control-point updates, yielding a continuously collision-free motion. 
    The generated free regions and candidate motions are further maintained in a region-based graph for incremental planning.
    }
    \label{fig:pipeline}
    \vspace{-12pt}
\end{figure*}

As illustrated in Fig.~\ref{fig:pipeline}, the proposed framework takes local LiDAR observations as input and produces continuously collision-free motions in four stages: gap-based navigation direction extraction (Section III-A), direction-aware free-region generation (Section III-B), continuous-safe trajectory generation (Section III-C), and region-based graph maintenance for incremental planning (Section III-D).

\subsection{Navigation Direction Extraction}

The input point cloud is projected onto a spherical range map parameterized by azimuth and elevation, which degenerates to a single-row angular range map in 2D, as illustrated in Fig.~\ref{fig:pipeline}(b).
Candidate gaps are detected as depth discontinuities between neighboring obstacle measurements, and representative directions are extracted from these angular openings for the subsequent region generation module.

\subsection{Direction-Aware Convex Free-Region Generation}

As illustrated in Fig.~\ref{fig:pipeline}(c), given a candidate navigation direction, we construct a convex collision-free polyhedron by iteratively generating separating hyperplanes between the robot and surrounding obstacles. 
The candidate direction biases both obstacle selection and hyperplane generation so that the resulting free region better preserves traversable space along that direction.

\paragraph{Direction representation}

Let $p_1 \in \mathbb{R}^d$ denote the current robot position and let $d \in \mathbb{R}^d$ be a candidate navigation direction extracted from the gap-based perception module. 
The unit vector
\[
e = \frac{d}{\|d\|}
\]
represents the desired motion direction and is used to guide obstacle selection and hyperplane generation.

\paragraph{QP-based hyperplane generation}

Obstacle points are processed iteratively to construct separating hyperplanes that expand the collision-free region. 
At each iteration, an obstacle point $o_i$ is selected using a direction-aware strategy. 
Obstacle points located in the forward half-space of the candidate direction are prioritized, and among them the point closest to the direction ray is selected. 
If no such point exists, the obstacle closest to the robot center is chosen.

For the selected obstacle point $o_i$, a separating hyperplane is computed by solving the following QP for the hyperplane normal $n$:
\begin{equation}
\begin{aligned}
\min_{n} \quad & n^\top Q n \\
\text{s.t.} \quad 
& (v_j - o_i)^\top n \le -\epsilon, \quad \forall j, \\
& (p_1 - o_i)^\top n = -1,
\end{aligned}
\end{equation}
where $v_j$ denotes the vertices of the convex robot body. 
The matrix $Q$ introduces a directional bias according to the candidate navigation direction,
\[
Q = ee^\top + \lambda (I-ee^\top),
\]
which encourages the hyperplane normal to align with $e$, resulting in a hyperplane approximately perpendicular to the desired motion direction. 
In this way, the generated polyhedron tends to preserve more free space along the candidate direction while still separating the robot from surrounding obstacles. 
For obstacles located behind the robot, the weighting controlled by $\lambda$ allows the hyperplane orientation to rotate so that the backward space can be truncated more efficiently.
The inequality constraints keep all robot vertices on the safe side of the hyperplane, while the equality constraint fixes the normal vector's scale and keeps the robot center inside the safe region.

The generated hyperplane is added to the polyhedral representation, and obstacle points lying outside the new half-space are removed from further consideration. 
This process is repeated until all obstacle points have been separated. 
The intersection of all generated half-spaces defines the final convex collision-free polyhedron. 





\subsection{Safe Trajectory Generation}

Once a convex collision-free polyhedron has been constructed, the next step is to generate a feasible trajectory that remains inside the free region while making progress along the navigation direction.

\paragraph{Target pose selection}

Let the convex free region be represented as the intersection of half-spaces
\[
\mathcal{P} = \{p \in \mathbb{R}^d \mid A p \le b\},
\]
where each row of $A$ corresponds to the normal vector of a supporting hyperplane and $b$ defines the corresponding offsets. 
Here, $p$ denotes the position of the robot reference point in the workspace.
Given the navigation direction $e$, the target pose is selected by sampling robot orientations and, for each sampled orientation, solving for the farthest position that the robot can reach along $e$ while remaining inside the free region.

Let $R$ denote a sampled robot orientation. 
In the planar case, $R$ is determined by a yaw angle, while in the spatial case, it is determined by sampled roll-pitch-yaw angles. 
For a convex robot with body vertices $v_j$, the occupancy of the oriented robot at position $p$ is constrained by
\[
a_k^\top p + \max_j a_k^\top (R v_j) \le b_k, \quad \forall k,
\]
where $a_k^\top x \le b_k$ denotes the $k$-th half-space of the polyhedron. 
For a fixed orientation $R$, the farthest reachable position along the navigation direction is then obtained by

\[
\begin{aligned}
p^\star(R) = 
\arg\max_{p} \quad & e^\top p \\
\text{s.t.} \quad 
& a_k^\top p + \max_j a_k^\top (R v_j) \le b_k,\ \forall k .
\end{aligned}
\]

This yields a candidate target pose for the sampled orientation. 
By evaluating all sampled orientations, we obtain a set of feasible target poses and select the one that achieves the largest forward progress along $e$. 
This sampled-orientation strategy avoids a more expensive joint nonconvex search over position and orientation while remaining effective for online local planning.

\paragraph{Trajectory parameterization}

Given the selected target pose, we parameterize the robot motion using Bézier curves for both position and orientation:
\[
p(t)=\sum_{i=0}^{k} B_i^k(t) P_i,\quad
\theta(t)=\sum_{i=0}^{k} B_i^k(t)\Theta_i,\quad t\in[0,1],
\]
where $P_i$ and $\Theta_i$ are the control points for position and orientation, respectively, and $B_i^k(t)$ are the Bernstein basis polynomials. 
The initial and terminal control points are fixed by the current robot pose and the target pose, while the remaining control points are iteratively adjusted to improve smoothness and enforce safety.

\paragraph{Continuous safety verification}

Let a supporting hyperplane of the convex free region be defined as
\[
a^\top x \le b .
\]
To guarantee collision-free motion, the entire robot body must remain inside this half-space throughout the trajectory. 
Let $v_j$ denote the robot vertices in the body frame, and let $R(t)$ be the rotation matrix determined by the orientation trajectory $\theta(t)$. 
Then the collision-free condition can be written as
\[
a^\top p(t) + \max_j a^\top\!\big(R(t)v_j\big) \le b .
\]

Define the violation function
\[
g(t)=a^\top p(t)+\max_j a^\top\!\big(R(t)v_j\big)-b .
\]
Ensuring collision-free motion is equivalent to verifying $g(t)\le 0$ for all $t\in[0,1]$.
Since $p(t)$ and $\theta(t)$ are smooth Bézier curves, $g(t)$ is Lipschitz continuous. 
Let
\[
r_{\max}=\max_j \|v_j\|,\quad
a_{\max}=\max_k \|a_k\|,
\]
and let $\bar v_p$ and $\bar v_\theta$ be upper bounds on the translational and rotational rates, obtained from the Bézier control-point differences. 
By upper-bounding the translational and rotational contributions to the rate of the support function, we obtain a valid global Lipschitz constant
\[
L = a_{\max}\big(\bar v_p + r_{\max}\bar v_\theta\big).
\]

For any interval $[t_L,t_R]$, let $t_c=(t_L+t_R)/2$. 
Using the Lipschitz property, we define the upper bound
\[
u(t_L,t_R) := g(t_c)+L\frac{t_R-t_L}{2},
\]
which satisfies
\[
\max_{t\in[t_L,t_R]} g(t) \le u(t_L,t_R).
\]
Thus, the interval is certified safe if $u(t_L,t_R)\le 0$.

\begin{figure}[t]
    \centering
    \includegraphics[width=\linewidth]{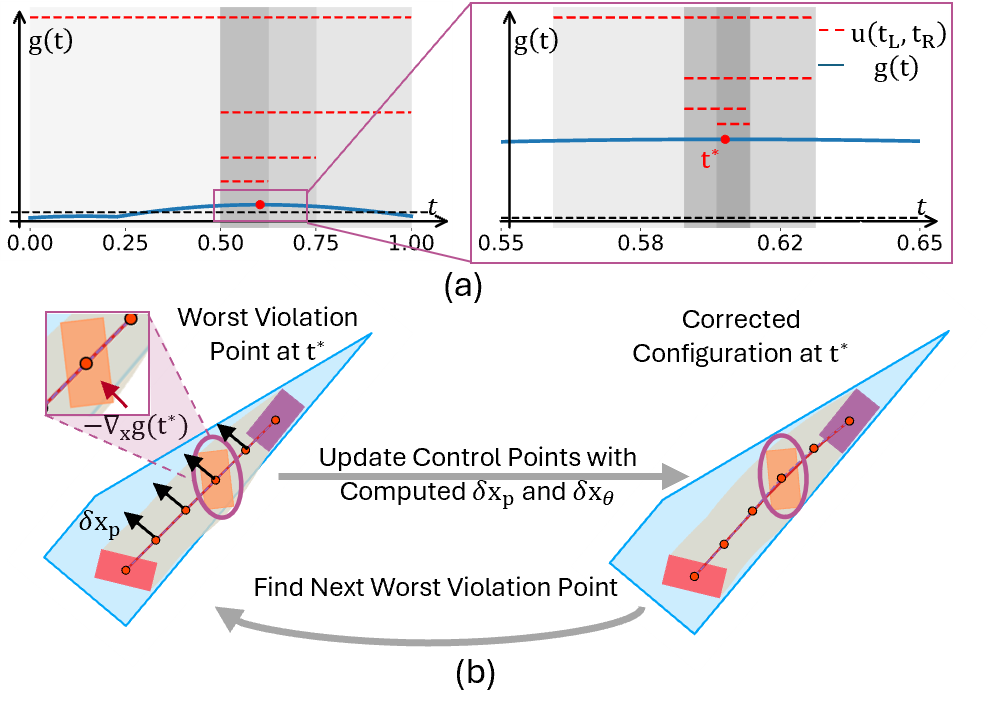}
    \caption{
    Safe trajectory generation via Lipschitz-based verification and local refinement.
    (a) The violation function $g(t)$ measures the signed distance to the supporting hyperplanes, where $g(t)\le 0$ indicates safety. 
    Lipschitz upper bounds $u(t_L,t_R)$ enable adaptive subdivision to localize the most critical time $t^\star$. 
    (b) At $t^\star$, the gradient $\nabla g(t^\star)$ indicates the ascent direction of violation, and the control points are updated via $\delta x$ in the descent direction to enforce $g(t^\star)\le 0$.
    }
    \label{fig:traj_refinement}
    \vspace{-0.5cm}
\end{figure}

Based on this bound, we perform adaptive interval subdivision to search for the worst constraint violation along the trajectory. 
Intervals with $u(t_L,t_R)\le 0$ are safely discarded, while the remaining intervals are selectively subdivided, focusing on the most critical portions of the trajectory. 
The process terminates when either all intervals are certified safe or a most critical time instant $t^\star$ is identified for further refinement. 
As illustrated in Fig.~\ref{fig:traj_refinement}(a), this procedure both provides a safety certificate and localizes the time instant $t^\star$ corresponding to the highest potential violation, which is subsequently used to guide local trajectory refinement.

\paragraph{Local trajectory refinement}

When a positive worst violation is detected, the trajectory is locally refined around the corresponding time instant $t^\star$. 
As shown in Fig.~\ref{fig:traj_refinement}(b), we exploit the local structure of the violation function to efficiently correct the trajectory.
At $t^\star$, we first identify the most violated supporting hyperplanes and the corresponding active robot vertices. 
We then select a small subset of internal Bézier control points with the largest Bernstein weights at $t^\star$, since these control points have the strongest local influence on the trajectory near the violating time.

Let $\delta x$ denote the stacked update vector of the selected position and orientation control points, and let $s\ge 0$ be a slack variable. 
Using a first-order approximation of the violation function at $t^\star$, the refinement step is formulated as a QP
\begin{equation}
\begin{aligned}
\min_{\delta x,\, s}\quad & \|\delta x_p\|_{W_p}^2 + \|\delta x_\theta\|_{W_\theta}^2 + w_s s^2 \\
\text{s.t.}\quad 
& \nabla g_\ell(t^\star)^\top \delta x - s \le -g_\ell(t^\star)-m,
\quad \forall \ell, \\
& \delta x \in \mathcal{B}, \quad s \ge 0,
\end{aligned}
\end{equation}
where $g_\ell(t^\star)$ denotes the violation of the $\ell$-th selected hyperplane at $t^\star$, $m>0$ is a safety margin, and $\mathcal{B}$ is a trust region that bounds the control-point updates. 
The weights $W_p$ and $W_\theta$ penalize translational and rotational changes, respectively.

After solving the QP, the resulting control-point update is applied with a backtracking line search. 
The updated trajectory is accepted if the worst continuous violation decreases; otherwise, the step size is reduced. 
This verification-and-refinement process is repeated until the trajectory becomes continuously safe, no further improvement can be found, or a predefined maximum number of iterations (set to 30 in our implementation) is reached, which guarantees termination of the procedure.
The overall safe trajectory generation procedure is summarized in Algorithm~\ref{alg:safe_traj_generation}.

\begin{algorithm}[t]
\caption{Trajectory Generation with Continuous Verification and Local Refinement}
\label{alg:safe_traj_generation}
\begin{algorithmic}[1]
\STATE \textbf{Input:} convex polyhedron $\mathcal{P}$, navigation direction $e$, initial pose $(p_0,\theta_0)$, maximum iterations $N$
\STATE \textbf{Output:} continuously safe B\'ezier trajectory $(p(t),\theta(t))$
\STATE Sample robot orientations
\FOR{each sampled orientation $R$}
    \STATE Compute the farthest feasible target pose along $e$
\ENDFOR
\STATE Select a target pose $(p^\star,\theta^\star)$
\STATE Initialize B\'ezier trajectory from $(p_0,\theta_0)$ to $(p^\star,\theta^\star)$
\FOR{$k = 1$ to $N$}
    \STATE Perform continuous safety verification
    \IF{trajectory is certified safe}
        \STATE \textbf{break}
    \ENDIF
    \STATE Identify the worst violating time $t^\star$
    \STATE Collect the most violated hyperplanes at $t^\star$
    \STATE Select control points with the largest Bernstein weights at $t^\star$
    \STATE Solve the local refinement QP
    \STATE Apply the control-point update with line search
\ENDFOR
\end{algorithmic}
\end{algorithm}
\vspace{-0.5cm}

\subsection{Navigation Graph}


The generated free regions and candidate motions are maintained in a region-based navigation graph for incremental planning in unknown environments.
Each node stores a robot pose and a locally constructed convex free region, and each edge represents a candidate motion from that region to a target pose along an extracted direction.
At each planning step, the current node is expanded into multiple frontier edges, and the edge whose target pose is closest to the global goal is selected for execution.
Trajectory generation with continuous safety verification is performed only for the selected edge.
If the edge is continuously safe, the robot executes it and adds a new node at the reached target pose.
Otherwise, the edge is marked invalid, and the planner selects the next most promising frontier edge, enabling recovery from locally unproductive branches through previously generated free regions.
\section{EXPERIMENTAL RESULTS}

In this section, we evaluate the proposed framework through validation of direction-aware free-region generation, quantitative studies in cluttered 2D navigation, and additional demonstrations in 3D and real-world environments. 

\subsection{Validation of the Proposed Direction-Aware Free-Region Generation}


We validate the proposed direction-aware free-region generation method in narrow-passage scenarios, where region quality depends not only on collision-free geometry but also on whether it supports traversal along a feasible direction.
Existing methods, such as RILS~\cite{SikangLiu}, FRTree~\cite{FRTree}, and FIRI~\cite{GaoFeiFreeSpace}, mainly construct convex regions from the local obstacle configuration around the seed, without explicitly considering navigation directions.

\begin{figure}[tbp]
    \centering
    \includegraphics[width = 0.47\textwidth]{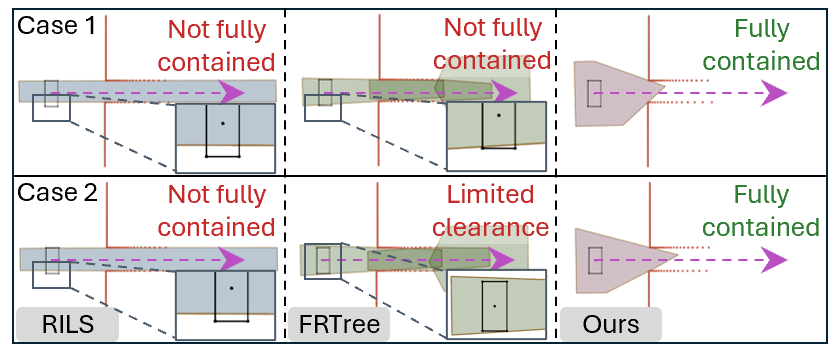}
    \vspace{-0.2cm}
    \caption{Free-region generation in two narrow-passage scenarios. 
    The black rectangle denotes the robot body, and the purple dashed arrow indicates the candidate direction. 
    With the same direction seed, RILS and FRTree may fail to fully contain the robot or leave limited clearance, while ours maintains full containment.}
    \label{fig:freespace}
    \vspace{-0.3cm}
\end{figure}


Fig.~\ref{fig:freespace} shows two representative examples.
RILS and FRTree use seed lines along the same candidate direction as ours.
RILS may produce slender regions that do not fully contain the robot body, while FRTree improves coverage through overlapping local inflations but can still leave insufficient clearance.
FIRI also focuses on local free-space enlargement without explicitly considering the candidate direction and is not shown because its implementation is unavailable.
In contrast, our method fully contains the robot body while preserving traversable space along the candidate direction.

\subsection{Simulation}

Before presenting the comparative simulations, we report the runtime of the main modules.
Table~\ref{tab:module_time} shows the average runtime of free-region generation, target pose selection, and safe trajectory generation in 2D and 3D settings, indicating that the proposed pipeline remains computationally efficient.

\begin{table}[tbp]
    \centering
    \caption{Average Runtime of Main Modules in 2D/3D Settings.}
    \label{tab:module_time}
    \setlength{\tabcolsep}{6pt}
    \begin{tabular}{c c c}
        \toprule
        Setting & Module & Avg. Time (ms) \\
        \midrule
        \multirow{3}{*}{2D}
        & Free-Region Generation      & 1.488 \\
        & Target Pose Selection      & 1.328 \\
        & Safe Trajectory Generation & 1.628 \\
        \midrule
        \multirow{3}{*}{3D}
        & Free-Region Generation      & 2.057 \\
        & Target Pose Selection      & 16.816 \\
        & Safe Trajectory Generation & 3.473 \\
        \bottomrule
    \end{tabular}
    \vspace{-0.4cm}
\end{table}


\subsubsection{Navigation in Cluttered and Unknown 2D Environments}

\begin{figure}[tbp]
    \centering
    \includegraphics[width = 0.48\textwidth]{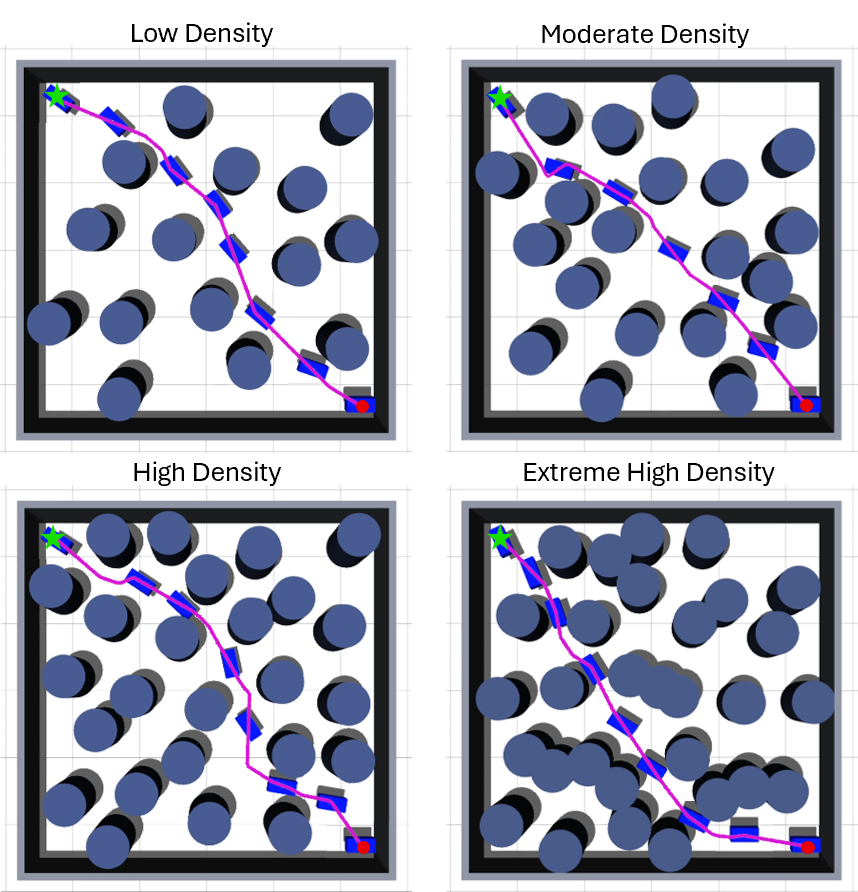}
    \caption{
    Representative 2D environments with obstacle densities of $0.6$, $0.8$, $1.0$, and $1.2~\mathrm{obstacles}/\mathrm{m}^2$. Each $5\,\mathrm{m} \times 5\,\mathrm{m}$ workspace contains cylindrical obstacles of radius $0.3\,\mathrm{m}$. Trajectories generated by the proposed framework are shown, illustrating safe navigation in cluttered scenes of varying density.}
    \label{fig:2d_simulation}
    \vspace{-0.3cm}
\end{figure}

We compare the proposed framework with DDR-OPT~\cite{ESDF_UniversalTO}, FASTER~\cite{faster}, and FRTree~\cite{FRTree} in randomly generated 2D scenes.
Each scene is a $5\,\mathrm{m} \times 5\,\mathrm{m}$ workspace with cylindrical obstacles of radius $0.3\,\mathrm{m}$, under obstacle densities of $0.6$, $0.8$, $1.0$, and $1.2~\mathrm{obstacles}/\mathrm{m}^2$ (Fig.~\ref{fig:2d_simulation}).
For each density, we generate five scenarios and conduct four trials per scenario, yielding 20 trials per density per method; results are summarized in Table~\ref{tab:2d_comparison}.
All methods use the same scenarios, start--goal pairs, robot footprint, and success/collision criteria.
DDR-OPT uses a two-disc robot model, with map resolution, collision-checking resolution, and inflation parameters tuned to the robot size and obstacle radius.
FASTER and FRTree use the same robot footprint, with corresponding safety margins and region parameters tuned accordingly.

We report three metrics: \emph{Length Scale}, the ratio of path length to the straight-line start--goal distance; \emph{Complete Rate}, the fraction of trials reaching the goal; and \emph{Collision-Free Rate}, the fraction of trials without collision during execution, regardless of goal completion.

As shown in Table~\ref{tab:2d_comparison}, all methods complete the task in sparse scenes, but safety differences appear as density increases.
DDR-OPT and FASTER drop in both completion and collision-free rates, while FRTree maintains high completion but exhibits increasing collisions.
In contrast, the proposed method maintains full completion and collision-free navigation across all densities.

These differences are related to how the compared methods represent free space and enforce safety. DDR-OPT and FASTER do not explicitly model robot geometry during safety checking, which limits their reliability in dense clutter. 
FRTree models robot geometry explicitly but verifies safety only at sampled states, which can miss inter-sample collisions. 
The proposed framework incorporates feasible traversal directions during free-region generation while providing continuous safety guarantees, allowing it to better preserve passable narrow gaps. 
Although some baselines achieve smaller length scales in certain settings, these shorter paths are accompanied by collisions or reduced completion rates.

\begin{table}[tbp]
    \centering
    \caption{Comparison of key metrics for different planners in 2D environments with varying obstacle densities.}
    \label{tab:2d_comparison}
    \setlength{\tabcolsep}{5.2pt}
    \begin{tabular}{c c c c c c}
        \toprule
        \makecell{Obstacle\\Density} & Metric & \makecell{DDR-\\OPT} & FASTER & FRTree & Ours \\
        \midrule
        \multirow{3}{*}{0.6}
        & Length Scale          & 1.02 & 1.03 & 1.29 & 1.03 \\
        & Complete Rate         & 1.00 & 1.00 & 1.00 & 1.00 \\
        & Collision-Free Rate   & 1.00 & 1.00 & 1.00 & 1.00 \\
        \midrule
        \multirow{3}{*}{0.8}
        & Length Scale          & 1.09 & 1.10 & 1.22 & 1.12 \\
        & Complete Rate         & 1.00 & 0.90 & 1.00 & \textbf{1.00} \\
        & Collision-Free Rate   & 0.55 & 0.20 & 1.00 & \textbf{1.00} \\
        \midrule
        \multirow{3}{*}{1.0}
        & Length Scale          & 1.14 & 1.08 & 1.22 & 1.27 \\
        & Complete Rate         & 0.90 & 0.25 & 1.00 & \textbf{1.00} \\
        & Collision-Free Rate   & 0.45 & 0.15 & 0.60 & \textbf{1.00} \\
        \midrule
        \multirow{3}{*}{1.2}
        & Length Scale          & 1.13 & N/A  & 1.21 & \textbf{1.10} \\
        & Complete Rate         & 0.45 & 0.00 & 0.90 & \textbf{1.00} \\
        & Collision-Free Rate   & 0.35 & N/A  & 0.40 & \textbf{1.00} \\
        \bottomrule
    \end{tabular}
    \vspace{-0.5cm}
\end{table}

\subsubsection{Ablation Study}
To isolate the two main components, we evaluate two ablation variants at densities 1.0 and 1.2 using the same scenarios.
\textit{Ours w/o Direction-Aware} sets $Q$ to the identity matrix and disables direction-aware obstacle selection, reducing the region generation to standard separating hyperplane construction.
\textit{Ours w/o Continuous Safety} removes continuous safety verification and local refinement, replacing them with sampled safety checking; the two are removed jointly because refinement relies on the violation time and gradient from continuous verification.

As shown in Table~\ref{tab:ablation}, removing direction-aware region generation reduces the completion rate while the collision-free rate remains 1.00, indicating that the generated regions become poorly aligned with feasible traversal directions, preventing the robot from finding viable paths through narrow passages.
Removing continuous safety also reduces the completion rate, because without local refinement the planner must discard trajectories with even small violations rather than repairing them, and occasionally fails to detect inter-sample collisions.
Both ablation variants also exhibit larger length scales than the full method, indicating less efficient paths even in successful trials.
The full method achieves the best performance across all metrics, confirming that the two components address complementary aspects of cluttered navigation.

\begin{table}[tbp]
    \centering
    \caption{Ablation study in high-density scenes. LS: Length Scale. CR: Complete Rate. CFR: Collision-Free Rate.}
    \label{tab:ablation}
    \setlength{\tabcolsep}{3.5pt}
    \begin{tabular}{l c c c c c c}
        \toprule
        & \multicolumn{3}{c}{Density 1.0} & \multicolumn{3}{c}{Density 1.2} \\
        \cmidrule(lr){2-4} \cmidrule(lr){5-7}
        Variant & LS & CR & CFR & LS & CR & CFR \\
        \midrule
        Ours w/o Direction-Aware      & 1.61 & 0.80 & 1.00 & 1.20 & 0.75 & 1.00 \\
        Ours w/o Continuous Safety    & 1.58 & 0.95 & 0.95 & 1.19 & 0.70 & 0.85 \\
        Ours (Full)                   & \textbf{1.27} & \textbf{1.00} & \textbf{1.00} & \textbf{1.10} & \textbf{1.00} & \textbf{1.00} \\
        \bottomrule
    \end{tabular}
    \vspace{-0.3cm}
\end{table}

\begin{figure}[tp]
    \centering
    \includegraphics[width = 0.48\textwidth]{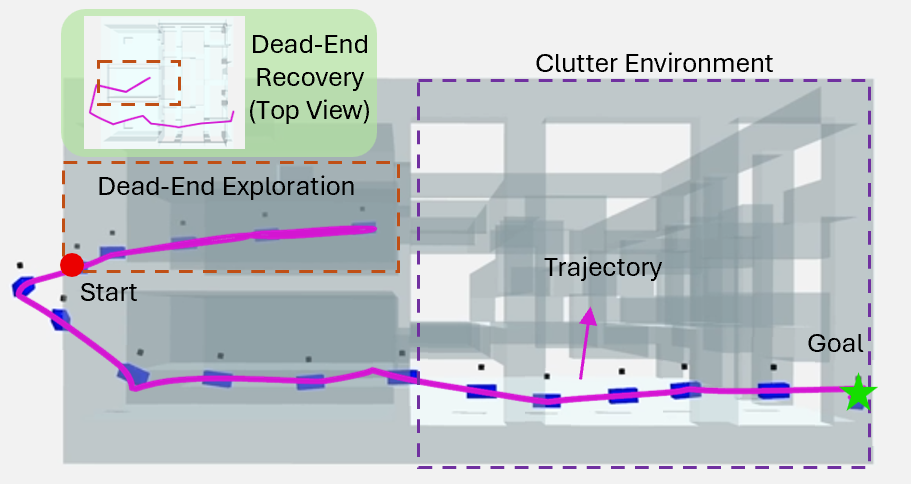}
    \caption{Representative result of 3D navigation in a cluttered environment with dead-end branches. 
    The figure shows that the proposed framework can generate collision-free motion in 3D and continue navigation after exploring a dead-end branch. 
    The upper-left top view highlights the corresponding backtracking behavior enabled by the maintained free-region graph.}
    \label{fig:3d_simulation}
    \vspace{-0.5cm}
\end{figure}

\subsubsection{Navigation in 3D Environments with Dead Ends}

We further provide a qualitative 3D experiment in a cluttered environment with dead-end branches to demonstrate the applicability of the proposed framework beyond planar navigation. In this experiment, the robot plans directly in 3D free space and navigates through the cluttered environment toward the goal.

As shown in Fig.~\ref{fig:3d_simulation}, the proposed method successfully guides the robot from the start to the goal while maintaining collision-free motion throughout the execution. 
The example also highlights the role of the region-based graph in handling dead ends. 
After entering a dead-end branch on the left side of the environment, the planner backtracks using the maintained graph of previously generated free regions and then continues toward the goal through another branch, as further illustrated in the top-view inset.


\subsection{Real-World Validation}

We finally validate the proposed framework in real-world experiments on two robotic platforms: a quadrupedal robot for 2D navigation and a UAV for 3D navigation.
This part is mainly intended to verify the practical applicability of the complete system in cluttered environments.

\begin{figure}[tbp]
    \centering
    \includegraphics[width = 0.48\textwidth]{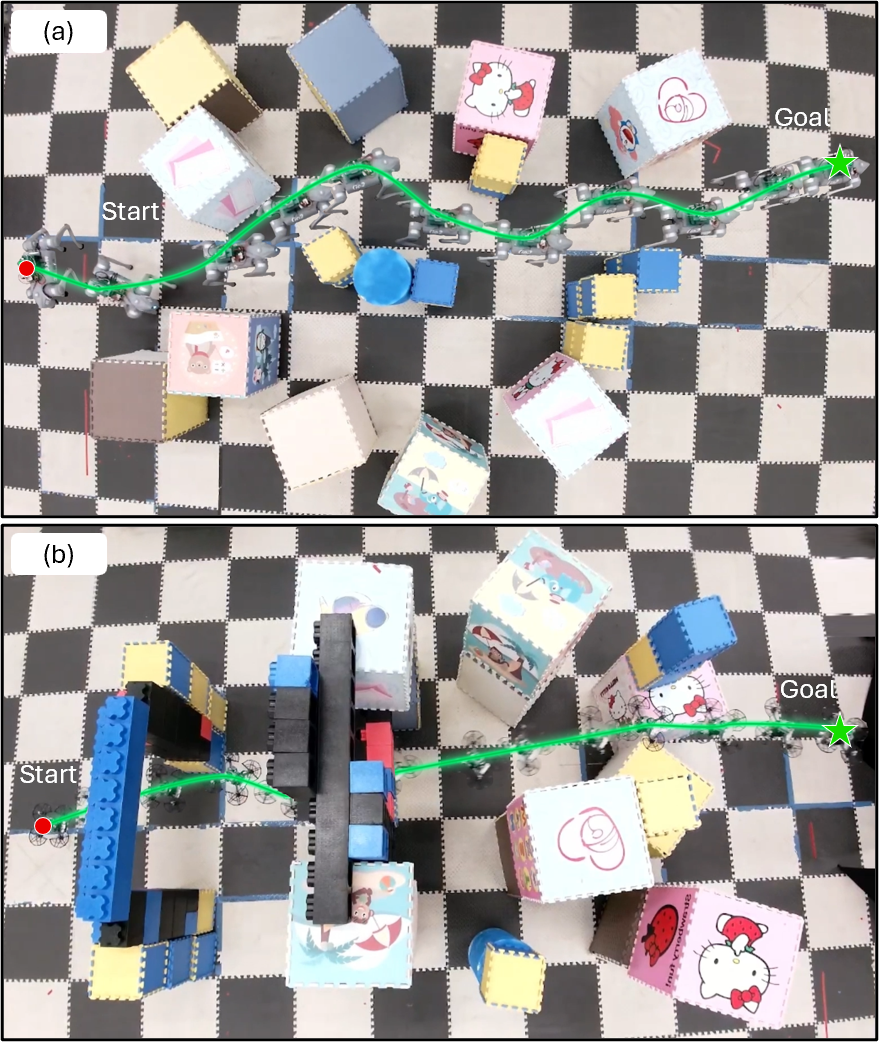}
    \caption{Real-world navigation results of the proposed framework on a quadrupedal robot and a UAV.
    (a) 2D navigation in a cluttered environment. 
    The robot successfully reaches the goal through narrow and cluttered regions without collision. 
    (b) 3D navigation in a cluttered environment with 3D obstacles of varying heights.}
    \label{fig:real_world_results}
    \vspace{-0.6cm}
\end{figure}

\subsubsection{2D Real-World Navigation}

We first test the proposed framework on a Unitree Go2 robot equipped with a MID360 LiDAR in an unknown and cluttered indoor environment. 
The robot is commanded to reach a designated goal in a $6\,\mathrm{m} \times 4\,\mathrm{m}$ workspace containing multiple obstacles. 
As shown in Fig.~\ref{fig:real_world_results}(a), the robot successfully navigates through cluttered regions and reaches the goal without collision. 

\subsubsection{3D Real-World Navigation}

We further evaluate the proposed framework on a UAV in a real-world 3D navigation scenario within a $6\,\mathrm{m} \times 4\,\mathrm{m} \times 2.5\,\mathrm{m}$ workspace containing obstacles of varying heights. 
As shown in Fig.~\ref{fig:real_world_results}(b), the robot successfully traverses the cluttered environment and reaches the goal without collision. 

\section{Discussion and Conclusion}
In this paper, we propose a robot navigation framework that integrates direction-aware convex free-region generation and continuous-safe trajectory generation for navigation in unknown and cluttered environments.
The proposed method produces free regions that better preserve traversable space along candidate directions, and generates continuously safe motion via Bezier trajectory parameterization, Lipschitz-based verification, and local control-point refinement. 
Experiments validate the framework in cluttered 2D navigation, with additional 3D and real-world demonstrations.

The proposed framework has several limitations that suggest directions for future work.
First, the framework does not guarantee completeness: in extremely elongated free regions, the fixed-degree Bézier parameterization may fail to produce a continuously safe trajectory even when a feasible path exists geometrically. 
This can be mitigated by tuning the $\lambda$ parameter in region generation or by increasing the number of Bézier control points at the cost of additional computation. 
Second, the current formulation addresses geometric continuous safety and delegates dynamic feasibility to the downstream tracking controller, following the common hierarchical planning-control architecture.
Explicit velocity and acceleration constraints can be naturally integrated into the refinement QP as a straightforward extension, which we leave to future work.

\bibliographystyle{IEEEtran}
\bibliography{ref}

@ARTICLE{GaoFeiFreeSpace,
  author={Wang, Qianhao and Wang, Zhepei and Wang, Mingyang and Ji, Jialin and Han, Zhichao and Wu, Tianyue and Jin, Rui and Gao, Yuman and Xu, Chao and Gao, Fei},
  journal={IEEE Transactions on Robotics}, 
  title={Fast Iterative Region Inflation for Computing Large 2-D/3-D Convex Regions of Obstacle-Free Space}, 
  year={2025},
  volume={41},
  number={},
  pages={3223-3243},
  doi={10.1109/TRO.2025.3562482}}

@ARTICLE{SikangLiu,
  author={Liu, Sikang and Watterson, Michael and Mohta, Kartik and Sun, Ke and Bhattacharya, Subhrajit and Taylor, Camillo J. and Kumar, Vijay},
  journal={IEEE Robotics and Automation Letters}, 
  title={Planning Dynamically Feasible Trajectories for Quadrotors Using Safe Flight Corridors in 3-D Complex Environments}, 
  year={2017},
  volume={2},
  number={3},
  pages={1688-1695},
  doi={10.1109/LRA.2017.2663526}}

@ARTICLE{FRTree,
  author={Li, Yulin and Song, Zhicheng and Zheng, Chunxin and Bi, Zhihai and Chen, Kai and Wang, Michael Yu and Ma, Jun},
  journal={IEEE Robotics and Automation Letters}, 
  title={{FRTree} Planner: Robot Navigation in Cluttered and Unknown Environments With Tree of Free Regions}, 
  year={2025},
  volume={10},
  number={4},
  pages={3811-3818},
  doi={10.1109/LRA.2025.3544519}}

@ARTICLE{faster,
  author={Tordesillas, Jesus and Lopez, Brett T. and Everett, Michael and How, Jonathan P.},
  journal={IEEE Transactions on Robotics}, 
  title={{FASTER}: Fast and Safe Trajectory Planner for Navigation in Unknown Environments}, 
  year={2022},
  volume={38},
  number={2},
  pages={922-938},
  doi={10.1109/TRO.2021.3100142}}

@ARTICLE{Navigationin2DClutteredScenes,
  author={Qu, Yinsong and Li, Yunxiang and Guo, Yudong and Yi, Weiyun and Cui, Hanwen and Lv, Yuezu and Zhong, Shanlin},
  journal={IEEE Robotics and Automation Letters}, 
  title={Manageable Convex Region Iteration without Half-Plane Optimization for Robot Navigation in 2D Cluttered Scenes}, 
  year={2026},
  volume={},
  number={},
  pages={1-8},
  doi={10.1109/LRA.2026.3668664}}

@ARTICLE{EllipsoidalRepresentation,
  author={Ruan, Sipu and Poblete, Karen L. and Wu, Hongtao and Ma, Qianli and Chirikjian, Gregory S.},
  journal={IEEE Transactions on Robotics}, 
  title={Efficient Path Planning in Narrow Passages for Robots With Ellipsoidal Components}, 
  year={2023},
  volume={39},
  number={1},
  pages={110-127},
  doi={10.1109/TRO.2022.3187818}}

@ARTICLE{Fast-Racing,
  author={Han, Zhichao and Wang, Zhepei and Pan, Neng and Lin, Yi and Xu, Chao and Gao, Fei},
  journal={IEEE Robotics and Automation Letters}, 
  title={Fast-Racing: An Open-Source Strong Baseline for $\mathrm{SE}(3)$ Planning in Autonomous Drone Racing}, 
  year={2021},
  volume={6},
  number={4},
  pages={8631-8638},
  doi={10.1109/LRA.2021.3113976}}

@ARTICLE{ylLocalReactive,
  author={Li, Yulin and Tang, Xindong and Chen, Kai and Zheng, Chunxin and Liu, Haichao and Ma, Jun},
  journal={IEEE Robotics and Automation Letters}, 
  title={Geometry-Aware Safety-Critical Local Reactive Controller for Robot Navigation in Unknown and Cluttered Environments}, 
  year={2024},
  volume={9},
  number={4},
  pages={3419-3426},
  doi={10.1109/LRA.2024.3360809}}

@ARTICLE{ylSOS,
  author={Li, Yulin and Zheng, Chunxin and Chen, Kai and Xie, Yusen and Tang, Xindong and Wang, Michael Yu and Ma, Jun},
  journal={IEEE Robotics and Automation Letters}, 
  title={Collision-Free Trajectory Optimization in Cluttered Environments Using Sums-of-Squares Programming}, 
  year={2024},
  volume={9},
  number={12},
  pages={11026-11033},
  doi={10.1109/LRA.2024.3486235}}

@INPROCEEDINGS{ESDF-Robo-Centric,
  author={Geng, Shuang and Wang, Qianhao and Xie, Lei and Xu, Chao and Cao, Yanjun and Gao, Fei},
  booktitle={2023 IEEE/RSJ International Conference on Intelligent Robots and Systems (IROS)}, 
  title={Robo-Centric ESDF: A Fast and Accurate Whole-Body Collision Evaluation Tool for Any-Shape Robotic Planning}, 
  year={2023},
  volume={},
  number={},
  pages={290-297},
  doi={10.1109/IROS55552.2023.10342074}}

@ARTICLE{ESDF_UniversalTO,
  author={Zhang, Mengke and Chen, Nanhe and Wang, Hu and Qiu, Jianxiong and Han, Zhichao and Ren, Qiuyu and Xu, Chao and Gao, Fei and Cao, Yanjun},
  journal={IEEE Transactions on Automation Science and Engineering}, 
  title={Universal Trajectory Optimization Framework for Differential Drive Robot Class}, 
  year={2025},
  volume={22},
  number={},
  pages={13030-13045},
  doi={10.1109/TASE.2025.3550676}}

@ARTICLE{hypersurface,
  author={Tordesillas, Jesus and How, Jonathan P.},
  journal={IEEE Transactions on Robotics}, 
  title={MADER: Trajectory Planner in Multiagent and Dynamic Environments}, 
  year={2022},
  volume={38},
  number={1},
  pages={463-476},
  doi={10.1109/TRO.2021.3080235}}

@article{syHypersurface,
  title={Online Trajectory Optimization for Arbitrary-Shaped Mobile Robots via Polynomial Separating Hypersurfaces},
  author={Li, Shuoye and Song, Zhiyuan and Li, Yulin and Bi, Zhihai and Ma, Jun},
  journal={arXiv preprint arXiv:2601.09231},
  year={2026}
}

@ARTICLE{3DFrontier,
  author={Batinovic, Ana and Petrovic, Tamara and Ivanovic, Antun and Petric, Frano and Bogdan, Stjepan},
  journal={IEEE Robotics and Automation Letters}, 
  title={A Multi-Resolution Frontier-Based Planner for Autonomous 3D Exploration}, 
  year={2021},
  volume={6},
  number={3},
  pages={4528-4535},
  doi={10.1109/LRA.2021.3068923}}

@INPROCEEDINGS{Frontier1997,
  author={Yamauchi, B.},
  booktitle={Proceedings 1997 IEEE International Symposium on Computational Intelligence in Robotics and Automation CIRA'97. 'Towards New Computational Principles for Robotics and Automation'}, 
  title={A frontier-based approach for autonomous exploration}, 
  year={1997},
  volume={},
  number={},
  pages={146-151},
  doi={10.1109/CIRA.1997.613851}}

@ARTICLE{2024RALFrontier,
  author={Zhang, Hong and Wang, Songyan and Liu, Yuanshuai and Ji, Pengtao and Yu, Runzhuo and Chao, Tao},
  journal={IEEE Robotics and Automation Letters}, 
  title={EFP: Efficient Frontier-Based Autonomous UAV Exploration Strategy for Unknown Environments}, 
  year={2024},
  volume={9},
  number={3},
  pages={2941-2948},
  doi={10.1109/LRA.2024.3363531}}

@INPROCEEDINGS{gapbased2010,
  author={Mujahad, Muhannad and Fischer, Dirk and Mertsching, Bärbel and Jaddu, Hussein},
  booktitle={2010 IEEE/RSJ International Conference on Intelligent Robots and Systems}, 
  title={Closest Gap based (CG) reactive obstacle avoidance Navigation for highly cluttered environments}, 
  year={2010},
  volume={},
  number={},
  pages={1805-1812},
  doi={10.1109/IROS.2010.5649736}}

@ARTICLE{potentialgap,
  author={Xu, Ruoyang and Feng, Shiyu and Vela, Patricio A.},
  journal={IEEE Robotics and Automation Letters}, 
  title={Potential Gap: A Gap-Informed Reactive Policy for Safe Hierarchical Navigation}, 
  year={2021},
  volume={6},
  number={4},
  pages={8325-8332},
  doi={10.1109/LRA.2021.3104623}}

@ARTICLE{safergap,
  author={Feng, Shiyu and Abuaish, Ahmad and Vela, Patricio A.},
  journal={IEEE Robotics and Automation Letters}, 
  title={Safer Gap: Safe Navigation of Planar Nonholonomic Robots With a Gap-Based Local Planner}, 
  year={2024},
  volume={9},
  number={12},
  pages={11034-11041},
  doi={10.1109/LRA.2024.3486231}}

@INPROCEEDINGS{sphere,
  author={Wu, Chengkai and Wang, Ruilin and Song, Mianzhi and Gao, Fei and Mei, Jie and Zhou, Boyu},
  booktitle={2024 IEEE International Conference on Robotics and Automation (ICRA)}, 
  title={Real-time Whole-body Motion Planning for Mobile Manipulators Using Environment-adaptive Search and Spatial-temporal Optimization}, 
  year={2024},
  volume={},
  number={},
  pages={1369-1375},
  doi={10.1109/ICRA57147.2024.10610192}}

@INPROCEEDINGS{kai,
  author={Chen, Kai and Liu, Haichao and Li, Yulin and Duan, Jianghua and Zhu, Lei and Ma, Jun},
  booktitle={2025 IEEE International Conference on Robotics and Automation (ICRA)}, 
  title={Robot Navigation in Unknown and Cluttered Workspace with Dynamical System Modulation in Starshaped Roadmap}, 
  year={2025},
  volume={},
  number={},
  pages={10140-10146},
  doi={10.1109/ICRA55743.2025.11128318}}

@inproceedings{RussDecompose,
  title={Computing large convex regions of obstacle-free space through semidefinite programming},
  author={Deits, Robin and Tedrake, Russ},
  booktitle={Algorithmic Foundations of Robotics XI: Selected Contributions of the Eleventh International Workshop on the Algorithmic Foundations of Robotics},
  pages={109--124},
  year={2015},
  organization={Springer}
}

@article{otte2016rrtx,
  title={RRTX: Asymptotically optimal single-query sampling-based motion planning with quick replanning},
  author={Otte, Michael and Frazzoli, Emilio},
  journal={The International Journal of Robotics Research},
  volume={35},
  number={7},
  pages={797--822},
  year={2016},
  publisher={SAGE Publications Sage UK: London, England}
}

@article{continuousCollisionCheck,
  title={Motion planning with sequential convex optimization and convex collision checking},
  author={Schulman, John and Duan, Yan and Ho, Jonathan and Lee, Alex and Awwal, Ibrahim and Bradlow, Henry and Pan, Jia and Patil, Sachin and Goldberg, Ken and Abbeel, Pieter},
  journal={The International Journal of Robotics Research},
  volume={33},
  number={9},
  pages={1251--1270},
  year={2014},
  publisher={Sage Publications Sage UK: London, England}
}

@inproceedings{merkt2019continuous,
  title={Continuous-time collision avoidance for trajectory optimization in dynamic environments},
  author={Merkt, Wolfgang and Ivan, Vladimir and Vijayakumar, Sethu},
  booktitle={2019 IEEE/RSJ International Conference on Intelligent Robots and Systems (IROS)},
  pages={7248--7255},
  year={2019},
  organization={IEEE}
}

@article{zhang2022generalized,
  title={A generalized continuous collision detection framework of polynomial trajectory for mobile robots in cluttered environments},
  author={Zhang, Zeqing and Zhang, Yinqiang and Han, Ruihua and Zhang, Liangjun and Pan, Jia},
  journal={IEEE Robotics and Automation Letters},
  volume={7},
  number={4},
  pages={9810--9817},
  year={2022},
  publisher={IEEE}
}

\end{document}